\documentclass[journal]{IEEEtran}

\ifCLASSOPTIONcompsoc
  \usepackage[nocompress]{cite}
\else
  \usepackage{cite}
\fi

\usepackage{amsfonts,dsfont}
\usepackage{balance}
\usepackage{booktabs}
\usepackage{amsmath}
\usepackage{tabularx}
\usepackage{subfigure}
\usepackage{graphicx}
\usepackage{multirow}
\usepackage{rotating}
\usepackage{algorithm}
\usepackage{algorithmic}
\usepackage{url}
\usepackage{amssymb}
\usepackage{amsthm}
\usepackage{enumerate}
\usepackage{epsfig}
\usepackage{epstopdf}
\usepackage{tikz}
\usepackage{adjustbox}
\usepackage{color}

\newtheorem{mythm}{Theorem}

\begin{document}
%
\title{Out-of-distribution Detection by Cross-class Vicinity Distribution of In-distribution Data}

\author{Zhilin Zhao,~
        Longbing Cao,~\IEEEmembership{Senior Member,~IEEE,}
        and~Kun-Yu Lin
\thanks{The work is partially sponsored by Australian Research Council Discovery and Future Fellowship grants (DP190101079 and FT190100734).}
\IEEEcompsocitemizethanks{\IEEEcompsocthanksitem Zhilin Zhao and Longbing Cao are with 
University of Technology Sydney, 
Australia. 
E-mail: zhi-lin.zhao@student.uts.edu.au, longbing.cao@gmail.com
\IEEEcompsocthanksitem Kun-Yu Lin is with 
Sun Yat-sen University, 
China. 
E-mail: kunyulin14@outlook.com
}
}

\markboth{}%
{\MakeLowercase{\textit{Zhao et al.}}: DA}

\maketitle

\begin{abstract}
Deep neural networks for image classification only learn to map in-distribution inputs to their corresponding ground truth labels in  training  without differentiating out-of-distribution samples from in-distribution ones. This results from the assumption that all samples are independent and identically distributed (IID) without distributional distinction. Therefore, a pretrained network learned from in-distribution samples treats out-of-distribution samples as in-distribution and makes high-confidence predictions on them in the test phase. To address this issue, we draw out-of-distribution samples from the vicinity distribution of training in-distribution samples for learning to reject the prediction on out-of-distribution inputs. A \textit{Cross-class Vicinity Distribution} is introduced by assuming that an out-of-distribution sample generated by mixing multiple in-distribution samples does not share the same classes of its constituents. We thus improve the discriminability of a pretrained network by finetuning it with out-of-distribution samples drawn from the cross-class vicinity distribution, where each out-of-distribution input corresponds to a complementary label. Experiments on various in-/out-of-distribution datasets show that the proposed method significantly outperforms the existing methods in improving the capacity of discriminating between in- and out-of-distribution samples.
\end{abstract}


%
\IEEEpeerreviewmaketitle

\section{Introduction}
\label{sec:introduction}

\begin{figure}
  \centering
  \includegraphics[width=1\linewidth]{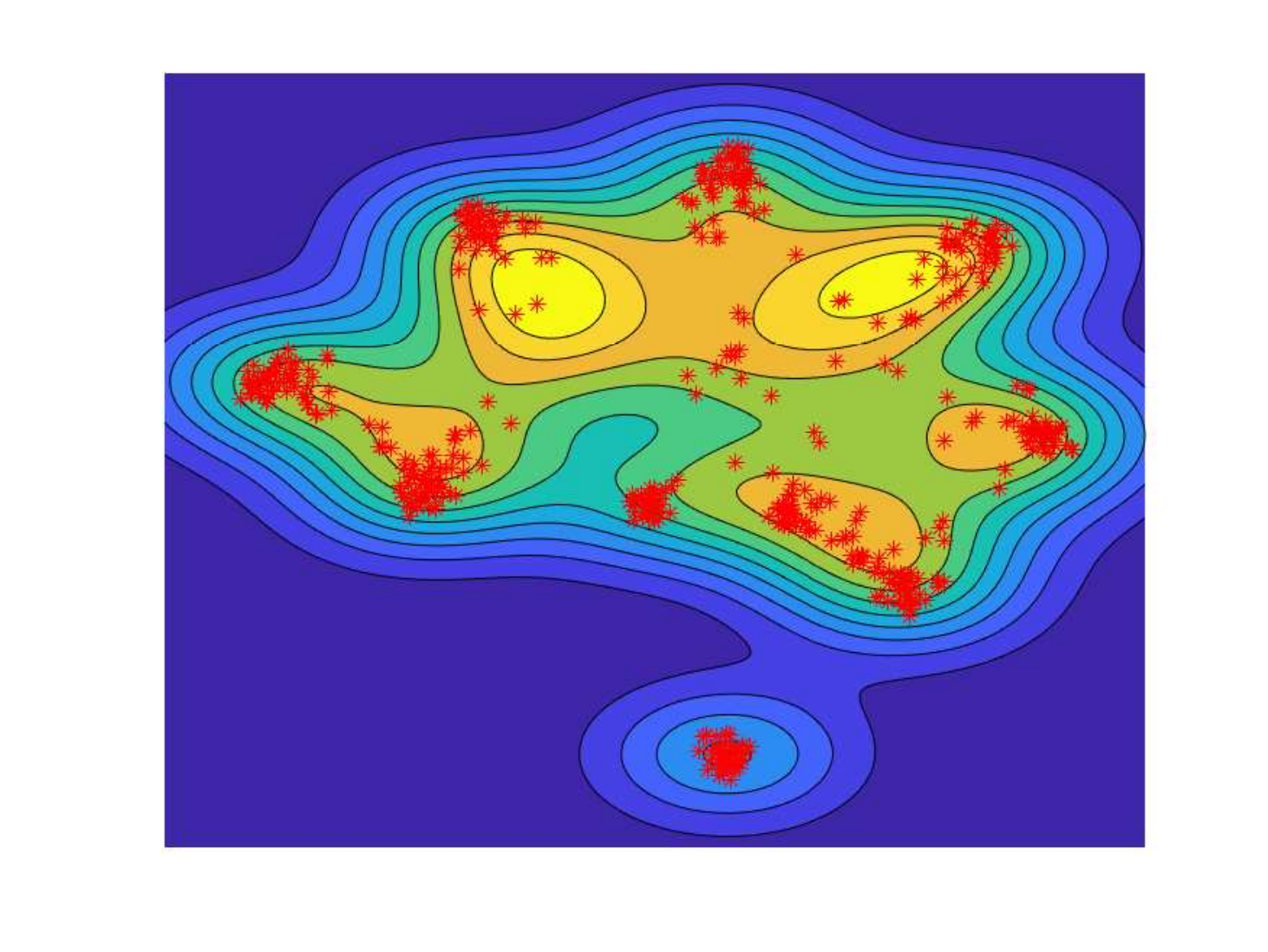}
  \caption{The heat map of prediction confidence by ResNet18 on CIFAR10. The embedding results are constructed by t-SNE~\cite{SNE:08}. Red points correspond to training in-distribution samples. Yellow regions correspond to high confidence for predictions, while blue regions correspond to low confidence. ResNet18 assigns high-confidence predictions on samples located in the regions outside the training in-distribution samples, i.e., out-of-distribution samples. It shows ResNet18 does not discriminate between in- and out-of-distribution samples. The figure is best viewed in color.}
  \label{fig:qt}
\end{figure}

\IEEEPARstart{A}{ccording} to the learning rule of empirical risk minimization~\cite{ML:14,SLT:98}, training a deep neural network for image classification is to minimize the average error over the independent and identically distributed (IID) samples drawn from an unknown distribution (called \textit{in-distribution})~\cite{NN:19}. The networks demonstrate a significant ability to recognize different classes of in-distribution samples and present a powerful generalization ability on test samples drawn from the same distribution~\cite{NN:12}. However, the test samples could be drawn from a distribution different from that of training in-distribution samples (called \textit{out-of-distribution}~\cite{BL:17}). The networks could assign unexpected high-confidence predictions on these out-of-distribution samples, as shown in \figurename~\ref{fig:qt}. The primary cause is that deep networks only map in-distribution inputs to the corresponding ground truth labels but never reject to map out-of-distribution inputs to any classes in the training phase. Therefore, a \textit{pretrained network}~\cite{UN:17} trained solely on in-distribution samples cannot effectively discriminate between in- and out-of-distribution samples, causing \textit{distributional vulnerability} \cite{Z22ood} and non-IID learning challenges \cite{C22beyiid}.

To address this issue in the pretrained network, the empirical risk minimization principle should be extended to consider out-of-distribution samples in the training process, i.e., going beyond the conventional IID assumption for training and test samplings. Specifically, one straightforward idea is to finetune the pretrained network with generated out-of-distribution samples if no extra real-world out-of-distribution samples are available~\cite{PN:18}. It is impossible to explore all out-of-distribution samples due to the infinite data space. Out-of-distribution samples are relative to an in-distribution dataset, e.g., an out-of-distribution sample for an in-distribution dataset could be in-distribution for other in-distribution datasets, and an in-distribution sample could be out-of-distribution for other in-distribution datasets. For finetuning a pretrained network to improve its out-of-distribution sensitivity, these out-of-distribution samples should be tailored for a given in-distribution dataset per its data characteristics, rejected to map to any classes and efficiently generated.

Accordingly, we construct a vicinity distribution~\cite{VRM:00} of training in-distribution samples to explore the corresponding out-of-distribution samples. For an in-distribution sample, we draw the related but different samples from its vicinity distributions, which is known as data augmentation~\cite{DA:96}. However, the standard construction of vicinity distributions~\cite{VRM:00,MIXUP:18} for the image classification task cannot generate data-dependent out-of-distribution samples. For an in-distribution sample, its horizontal reflection, rotation and rescaling can be made by the standard construction of vicinity distributions~\cite{DA:17}. These augmented samples are treated as in-distribution and mapped to the label as the original one to improve generalization. Without considering the vicinity relations across samples of different classes, these standard constructions assume that the samples from the same vicinity distribution own the same label and thus cannot generate out-of-distribution samples. Accordingly, we break down this barrier to construct a vicinity distribution of an in-distribution sample by involving the in-distribution samples of other classes.

For image classification by deep neural networks, we propose a \textit{Learning from Cross-class Vicinity Distribution} (LCVD) approach to explore out-of-distribution samples in relation to a given in-distribution sample according to the following insight:
\begin{quote}
\textit{An out-of-distribution input generated by mixing multiple in-distribution inputs does not belong to the same classes as its constituents.}
\end{quote}
To obtain an out-of-distribution sample drawn from the cross-class vicinity distribution of an in-distribution sample, we can linearly combine the in-distribution sample with multiple in-distribution samples corresponding to other classes. To encourage a network to make low-confidence predictions on samples located in the regions outside the training in-distribution samples, we assume the in-distribution sample no longer belongs to the original class after being contaminated by those in-distribution ones from other classes. This is because there is more than one class in the contaminated input (the generated out-of-distribution input). Accordingly, the generated out-of-distribution input corresponds to a \textit{complementary label}\footnote{Contrary to the definition of ground truth labels, complementary labels indicate the classes a given input does not belong to.}~\cite{CL:17} which could be any of the labels of its constituents. Because the generated out-of-distribution input does not belong to the label of its corresponding in-distribution input, we can finetune the pretrained network to reject the mapping between out-of-distribution inputs and the corresponding complementary labels. This improves the network capacity of discriminating between in- and out-of-distribution samples.

The main contributions of this work include the following:
\begin{enumerate}
\item According to the mutual information maximization, we derive a \textit{generic expected risk} for optimizing networks on in- and out-of-distribution samples.
\item Given an in-distribution dataset, we construct the \textit{cross-class vicinity distribution} to generate its corresponding out-of-distribution samples.
\item We improve the discriminability of a pretrained network by finetuning it with the generated out-of-distribution samples according to the \textit{generic empirical risk} of the generic expected risk.
\end{enumerate}

The rest of this paper is organized as follows. In Section~\ref{sec:RelatedWork}, we briefly review the related work. In Section~\ref{sec:proposedmethod}, we describe the proposed method in detail, including the generic expected risk, the vicinity distributions, and the generic empirical risk. Section~\ref{sec:pf} provides detailed proofs. In Section~\ref{sec:Experiments}, extensive experiments are conducted to verify the superiority of the proposed method. Section~\ref{sec:Conclusions} draws the conclusion and discusses the future work.

\section{Related Work}
\label{sec:RelatedWork}
The considered out-of-distribution detection task in this paper aims to detect whether a test sample for a network is in- or out-of-distribution in the test phase~\cite{GOOD:21, AOOD:21, GDNN:20}. The out-of-distribution detection problem is highly related to various research and literature fields, including predictive uncertainty~\cite{PU:17}, adversarial example detection~\cite{AD:15}, outlier detection~\cite{AD:21,OD:19}, and open set recognition~\cite{OSR:20,OSDA:21}.

In the out-of-distribution detection setup, the training samples are assumed to be in-distribution, and the test samples could be in- or out-of-distribution. For a network learned from in-distribution samples, out-of-distribution detection aims to detect the out-of-distribution samples with sematic shift~\cite{AOOD:21} according to the network outputs in the test phase. Similarly, predictive uncertainty and adversarial example detection detect some specific samples for a network learn from in-distribution samples in the test phase. Differently, predictive uncertainty~\cite{PU:17} mainly focuses on assigning high uncertainty to the in-distribution test samples that are misclassified by the network, while adversarial example detection~\cite{AD:15} concentrates on detecting the out-of-distribution samples with covariate shift that are applied for fooling the network. Note that out-of-distribution detection can only access in-distribution samples in the training phase and focus on detecting out-of-distribution samples for unlabeled test samples. Differently, outlier detection~\cite{AD:21,OD:19} identifies and filters out-of-distribution samples in the training dataset for learning downstream networks. Furthermore, open set recognition~\cite{OSR:20,OSDA:21} trains networks with unlabeled out-of-distribution samples and learns to assign them to extra unknown classes that do not belong to any classes of in-distribution samples. Specifically, the existing methods of out-of-distribution detection can be divided into three categories~\cite{GOOD:21}: post-hoc detection, confidence enhancement, and outlier exposure methods.

\subsection{Post-hoc Detection Methods}
Post-hoc detection methods aim to train out-of-distribution detectors for a pretrained network without modifying loss functions and learning processes. Specifically, out-of-distribution detectors calculate the out-of-distribution scores for test samples according to the outputs from networks and distinguish in- and out-of-distribution samples according to the scores. For example, the baseline method~\cite{BL:17} assumes out-of-distribution samples own a lower output confidence than in-distribution samples. Accordingly, the baseline method uses the confidence represented by maximum softmax outputs as the out-of-distribution scores for test samples. To make the confidence more sensitive to out-of-distribution, out-of-distribution detector for neural networks (ODIN)~\cite{ODIN:18} improves the baseline method by perturbing the inputs with negative adversarial direction~\cite{VAT:19,AD:15} and incorporating temperature scaling into the softmax function. Rather than applying the prediction confidence, MahaLanoBis (MLB)~\cite{MLB:18} calculates the Mahalanobis distances of output features for each network layer and integrates the Mahalanobis distances on all layers by training a logistic regression detector. Not only accessing the maximal prediction confidence, an energy-based detector~\cite{EB:20} uses the predicted label probabilities represented by softmax activation to calculate an energy function score. Rectified Activations (RA)~\cite{RA:21} truncates the activation on the penultimate layer of a network to decrease the negative effect of noise in calculating the confidence. For the Kullback-Leibler divergence between network softmax outputs and a uniform distribution, GradNorm~\cite{IG:21,ED:22} applies the norm of gradients with respect to weights as out-of-distribution scores. The out-of-distribution detectors learn the knowledge which is sensitive to out-of-distribution samples from the outputs of pretrained networks without refining the networks, which indicates that the detector performance relies heavily on the properties of the pretrained networks.

\subsection{Confidence Enhancement Methods}
To improve the out-of-distribution sensitivity of a pretrained network, confidence enhancement methods aim to retrain or finetune the pretrained network by modifying loss functions and learning processes. Based on the baseline method~\cite{BL:17}, Deep Gambler (DG)~\cite{DG:19} sets a threshold for networks in the training phase, and the network abstains from making a prediction when the confidence is lower than the predetermined threshold. Built on ODIN, DeConf-C~\cite{DCC:20} improves the traditional cross-entropy loss by decomposing confidence and modifies the learning process by searching adversarial perturbation magnitudes on training in-distribution samples. Based on the observation that a large number of classes leads to decreased out-of-distribution detection performance, Minimum Other Score (MOS)~\cite{MOS:21} explores the hierarchical structure of labels and divides the training in-distribution samples into several groups according to labeling concepts.

\subsection{Outlier Exposure Methods}
Outlier exposure methods retrain or finetune a pretrained network by involving out-of-distribution samples in the training process. They can be treated as specific confidence enhancement methods. The training out-of-distribution samples can be synthetic and obtained from real-world datasets. The basic Outlier Exposure (OE)~\cite{OE:19} method finetunes a pretrained network with a marginal loss to enlarge the confidence gap between training in- and out-of-distribution samples. Built on OE, Outlier Exposure with Confidence Control (OECC)~\cite{OEC:21} encourages the output distribution of out-of-distribution samples to be close to a uniform distribution and minimizes the Euclidean distance between training classification accuracy and its confidence. Posterior Sampling-based Outlier Mining (POEM)~\cite{POEM:22} focuses on finding near-boundary out-of-distribution samples to learn a compact decision boundary between in- and out-of-distribution samples.

However, the aforementioned outlier exposure methods require real-world out-of-distribution samples to be given, which may be unavailable in practice. Accordingly, some other outlier exposure methods fill this gap by synthesizing out-of-distribution samples. For example, Joint Confidence Loss (JCL)~\cite{GO:18} generates out-of-distribution samples by a generative adversarial network~\cite{GAN:14}, which finds the samples around the boundary of training in-distribution samples. JCL then encourages the label probability vectors of the generated out-of-distribution samples to satisfy a uniform distribution while learning to classify in-distribution samples. Conversely, Joint Energy-based Model (JEM)~\cite{EB:20} generates in-distribution samples from the implicit generator inferred from the logits of a network and encourages the generated and original in-distribution samples to have similar network outputs. Instead of generating samples from generators, MIXUP~\cite{MIXUP:18} efficiently generates new in-distribution samples by augmenting the existing in-distribution samples. Specifically, MIXUP convexly combines two randomly-selected in-distribution inputs and the corresponding labels to construct an out-of-distribution sample. Contrasting Shifted Instances (CSI)~\cite{CSI:20} and Self-Supervised outlier Detection (SSD)~\cite{SSD:21} treat the rotated version of an in-distribution sample as the positive one and the other in-distribution samples as negative in contrastive losses. CSI calculates out-of-distribution scores according to the cosine similarity to the nearest training sample and the norm of the representation, while SSD applies shrunk covariance estimators~\cite{SCE:04} and data augmentation. Furthermore, contrastive learning is applied to transformers to improve the out-of-distribution detection performance of text classification~\cite{PT:21}.

The outlier exposure methods, focusing on synthesizing out-of-distribution samples when these samples are unavailable, are most related to the proposed LCVD method. JCL and JEM apply extra generators for obtaining out-of-distribution samples specific for the training in-distribution dataset, which is expensive. Although the other methods are effective in generating out-of-distribution samples by augmenting in-distribution samples, the augmented samples do not directly encourage networks to make low-confidence predictions for out-of-distribution samples. Different from the existing methods of synthesizing out-of-distribution samples, the proposed LCVD method aims to effectively and directly construct out-of-distribution samples by augmenting in-distribution samples that are used to be refused by networks.

\section{The LCVD Method}
\label{sec:proposedmethod}
The proposed LCVD method aims to improve the ability of a pretrained network in discriminating between in- and out-of-distribution samples for the image classification task. This is achieved by finetuning it with the data-dependent out-of-distribution samples drawn from the cross-class vicinity distribution of training in-distribution samples. In this section, we describe the generic expected risk considering both in- and out-of-distribution samples in the learning process, the cross-class vicinity distribution for sampling out-of-distribution samples, and the generic empirical risk which is used to finetune the pretrained network.

\subsection{Generic Expected Risk}
Here, we introduce a generic expected risk~\cite{CO:08} for learning a network from both in- and out-of-distribution samples. We refer $\mathbf{x}$ and $y$ to the input and label, respectively, and the number of classes is $K$. Each in-distribution input corresponds to a ground truth label, and each out-of-distribution input corresponds to a complementary label. For in-distribution samples, we assume the marginal distribution of $\mathbf{x}$, the marginal distribution of $y$, and the joint distribution are $P_I(\mathbf{x})$, $P_I(y)$ and $P_I(\mathbf{x},y)$, respectively. Correspondingly, we assume $P_O(\mathbf{x})$, $P_O(y)$ and $P_O(\mathbf{x},y)$ represent out-of-distribution samples. We assume both in- and out-of-distribution samples share the same conditional distribution $P(y | \mathbf{x})$, which is estimated by a parameterized network $Q_{\theta}(y | \mathbf{x})$ with model parameter $\theta$. This assumption is mild because the two kinds of samples share the same label space and $P(y | \mathbf{x})$ depends on the given input $x$. Further, estimating $Q_{\theta}(y | \mathbf{x})$ for the two kinds of samples can recognize different classes of in-distribution inputs and discriminate between in- and out-of-distribution samples. Mutual information~\cite{MINE:18} is a quantity of measuring the relationship between random variables. The mutual information for in- and out-of-distribution sample are $\mathcal{I}_I(\mathbf{x};y)$ and $\mathcal{I}_O(\mathbf{x};y)$, respectively.

The pretrained network focuses on measuring the relationship between a random in-distribution input and the corresponding ground truth label in the pretraining phase. It then rejects out-of-distribution samples in the finetuning phase to improve the out-of-distribution sensitivity. According to the definitions of in- and out-of-distribution samples, networks should enhance $\mathcal{I}_I(\mathbf{x};y)$ and reduce $\mathcal{I}_O(\mathbf{x};y)$, respectively, to improve the discriminability. Accordingly, we have,
\begin{equation}
\begin{aligned}
\max \quad \mathcal{I}_I(\mathbf{x};y) - \mathcal{I}_O(\mathbf{x};y).
\label{eq:mi}
\end{aligned}
\end{equation}
To maximize Eq.~(\ref{eq:mi}) and introduce $Q_{\theta}(y | \mathbf{x})$ to estimate $P(y | \mathbf{x})$, we obtain the lower bound of $\mathcal{I}_I(\mathbf{x};y)$ and the the upper bound of $\mathcal{I}_O(\mathbf{x};y)$:
\begin{equation}
\begin{aligned}
\mathcal{I}_I(\mathbf{x};y) \geq & \int \log Q_{\theta}(y | \mathbf{x}) {\rm d} P_I(\mathbf{x},y) + H(P_I(y)),
\label{eq:mii}
\end{aligned}
\end{equation}
and
\begin{equation}
\begin{aligned}
\mathcal{I}_O(\mathbf{x};y) \leq & - \int \log (1 - Q_{\theta}(y | \mathbf{x})) {\rm d} P_O(\mathbf{x},y) \\
& + \mathbb{E}_{P_O(y) } \left[ P_I(y)\right]^{-1},\\
\label{eq:mio}
\end{aligned}
\end{equation}
where $H(\cdot)$ is the entropy. The derivations are presented in Section~\ref{App:LWB} and Section~\ref{App:UWB}, respectively. Substituting Eq.~\eqref{eq:mii} and Eq.~\eqref{eq:mio} into Eq.~\eqref{eq:mi} and ignoring the constant terms, we obtain the generic expected risk for learning from both in- and out-of-distribution samples,
\begin{equation}
\begin{aligned}
\mathcal{R}(\theta) = & - \int \log Q_{\theta}(y | \mathbf{x}) {\rm d} P_I(\mathbf{x},y) \\
& +  \int \log (1 - Q_{\theta}(y | \mathbf{x})) {\rm d} P_O(\mathbf{x},y).
\label{eq:ger}
\end{aligned}
\end{equation}
The first term in $\mathcal{R}(\theta)$ is the expected risk for learning a pretrained $Q_{\theta}(y | \mathbf{x})$ without considering the out-of-distribution sensitivity. We finetune $Q_{\theta}(y | \mathbf{x})$ to improve the ability in discriminating between in- and out-of-distribution samples according to $\mathcal{R}(\theta)$.

\subsection{Cross-class Vicinity Distribution}

\subsubsection{Empirical Distribution}
Optimizing $Q_{\theta}(y | \mathbf{x})$ by minimizing $\mathcal{R}(\theta)$ is intractable because we usually cannot obtain the analytic expressions for $P_I(\mathbf{x},y)$ and $P_O(\mathbf{x},y)$. We can only access a training in-distribution dataset $\mathcal{D}_I = \{(\mathbf{x}_i^I,y_i^I)\}_{i = 1}^{N_I}$, where each sample $(\mathbf{x}_i^I,y_i^I)$ is assumed to be IID drawn from the unknown $P_I(\mathbf{x},y)$, and $N_I$ is the number of in-distribution samples. According to the vicinal risk minimization principle~\cite{VRM:00}, we can approximate $\mathcal{R}(\theta)$ by replacing $P_I(\mathbf{x},y)$ and $P_O(\mathbf{x},y)$ with the corresponding empirical distributions $\widetilde{P}_I(\mathbf{x},y)$ and $\widetilde{P}_O(\mathbf{x},y)$.
$\widetilde{P}_I(\mathbf{x},y)$ and $\widetilde{P}_O(\mathbf{x},y)$ are constructed by the corresponding vicinity distributions $\mathcal{VI}$ and $\mathcal{VO}$ based on the training in-distribution dataset $\mathcal{D}_I$, respectively,
\begin{equation}
\begin{aligned}
& \widetilde{P}_I(\mathbf{x},y) = \frac{1}{N_I} \sum_{i = 1}^{N_I} \mathcal{VI}( \mathbf{x},y | \mathbf{x}_i^I,y_i^I), \\
& \widetilde{P}_O(\mathbf{x},y) = \frac{1}{N_I} \sum_{i = 1}^{N_I} \mathcal{VO}( \mathbf{x},y | \mathbf{x}_i^I,y_i^I).
\label{eq:v}
\end{aligned}
\end{equation}

To obtain the two empirical distributions (i.e., $\widetilde{P}_I(\mathbf{x},y)$ and $\widetilde{P}_O(\mathbf{x},y)$) for exploring in- and out-of-distribution samples, we have to define the corresponding \textit{vicinity distributions} (i.e., $\mathcal{VI}$ and $\mathcal{VO}$) measuring the probability of finding the \textit{virtual} input-label pairs in the vicinity based on a given in-distribution sample $(\mathbf{x}^I,y^I)$ drawn from $P_I(\mathbf{x}, y)$. Note that $\widetilde{P}_O(\mathbf{x},y)$ is an empirical distribution for generating out-of-distribution samples, which are also built on the in-distribution samples as $\widetilde{P}_I(\mathbf{x},y)$. This is because the samples outside the in-distribution dataset could be out-of-distribution, and the vicinity distribution finding the neighborhood around an in-distribution sample could explore the data-dependent out-of-distribution samples.

\subsubsection{Dirac Delta Vicinity Distribution}
In the pretraining phase, we only apply the in-distribution samples to train a network without exploring the samples outside the training in-distribution dataset. Therefore, the corresponding \textit{dirac delta vicinity distribution} for in-distribution samples is defined as
\begin{equation}
\mathcal{VI}( \mathbf{x},y | \mathbf{x}^I,y^I) = \delta(\mathbf{x} = \mathbf{x}^I, y = y^I),
\label{eq:vi}
\end{equation}
where $\delta$ is the Dirac delta function. However, the vicinity distribution $\mathcal{VI}(\mathbf{x},y | \mathbf{x}^I,y^I)$ cannot be applied to find samples different from the in-distribution samples. This causes the uncertain predictions for out-of-distribution samples and the unexpected high-confidence predictions for some of them.

\subsubsection{Cross-class Vicinity Distribution}
Accordingly, we construct another vicinity distribution $\mathcal{VO}( \mathbf{x},y | \mathbf{x}^I,y^I)$ to explore out-of-distribution samples by considering the vicinity relations among the in-distribution samples of different classes. We then teach a pretrained network to reject the out-of-distribution samples drawn from $\widetilde{P}_O(\mathbf{x},y)$, which improves the out-of-distribution sensitivity. To consider the vicinity relations across in-distribution samples of different classes, we combine different classes of in-distribution inputs to generate an out-of-distribution input. However, it is hard to determine the ground truth label for the out-of-distribution input because the input contains more than one label information. To improve the discriminability of in- and out-of-distribution samples, we restrict networks to provide low-confidence predictions for inputs containing different label information. Consequently, it would be easy to determine the complementary labels for the out-of-distribution input, i.e., the ground truth labels of in-distribution inputs. We construct the vicinity distribution of in-distribution samples for exploring out-of-distribution samples based on the insight: an out-of-distribution input generated by mixing multiple in-distribution inputs does not belong to the classes of these in-distribution inputs.

For a given in-distribution sample $(\mathbf{x}^I,y^I)$, we firstly generate an out-of-distribution input $\mathbf{x}^O$ by combining $\mathbf{x}^I$ with other $M - 1 (M \leq 1000)$ randomly selected in-distribution inputs $\{\mathbf{x}_1^I, \ldots, \mathbf{x}_{M - 1}^I\}$,
\begin{equation}
\mathbf{x}^O = \frac{1}{M} (\mathbf{x}^I + \sum_{i = 1}^{M - 1} \mathbf{x}_i^I),
\label{eq:xo}
\end{equation}
the corresponding label set of the $M$ selected samples is $C(M) = (\bigcup_{i = 1}^{M - 1} y_i^I) \cup y^I \subseteq [K]$. Recall that the number of classes of in-distribution samples is $K$. We assume the number of selected classes in $C(M)$ is $K_C$.

\begin{mythm}
\label{th:kc}
Assume the out-of-distribution sample $\mathbf{x}^O$ is constructed by linearly combining $M (M \leq 1000)$ randomly selected in-distribution samples. With a high probability, the number of selected classes $K_C$ is less than $K$, which indicates the label set $C(M)$ is nearly impossible to contain all the labels of in-distribution samples.
\end{mythm}

The detailed proof is represented in Section~\ref{App:proof}. According to Theorem~\ref{th:kc}, an out-of-distribution input $\mathbf{x}^O$ is constructed by integrating $M$ in-distribution inputs that belong to $K_C$ classes. $K_C \leq K$ is with a high probability, which indicates that $C(M)$, which is the complementary label set of $\mathbf{x}^O$, does not cover all the classes with a high probability. Based on the insight of this construction method, $\mathbf{x}^O$ does not belong to the $K_C$ classes. Accordingly, to improve the out-of-distribution sensitivity, a pretrained network can reject to map this out-of-distribution input $\mathbf{x}^O$ to any of the $K_C$ classes. For the constructed out-of-distribution input $\mathbf{x}^O$, the complementary label is randomly selected from the complementary label set $C(M)$ and should be rejected by the pretrained network,
\begin{equation}
y^O \sim C(M) = (\bigcup_{i = 1}^{M - 1} y_i^I) \cup y^I.
\label{eq:yo}
\end{equation}

From the vicinity distribution perspective, we draw the out-of-distribution sample $(\mathbf{x}^O,y^O)$ from an empirical distribution constructed by a vicinity distribution of the in-distribution samples $(\mathbf{x}^I,y^I)$, which considers the vicinity relations among the samples of different classes. According to the input Eq.~(\ref{eq:xo}) and the complementary label Eq.~(\ref{eq:yo}) of the out-of-distribution sample, we obtain the following \textit{cross-class vicinity distribution} for exploring out-of-distribution samples based on the given in-distribution sample $(\mathbf{x}^I, y^I)$.
\begin{equation}
\begin{aligned}
\mathcal{VO}( \mathbf{x},y | \mathbf{x}^I,y^I) =  \mathbb{E}_{\mathbf{x}_1^I}\ldots\mathbb{E}_{\mathbf{x}_{M - 1}^I}\left[ \delta \left(\mathbf{x} = \mathbf{x}^O, y = y^O\right)\right],
\label{eq:vo}
\end{aligned}
\end{equation}
where $\mathbf{x}^O$ and $y^O$ are constructed by Eq~(\ref{eq:xo}) and Eq~(\ref{eq:yo}), respectively.

We can obtain the two empirical distributions $\widetilde{P}_I(\mathbf{x},y)$ and $\widetilde{P}_O(\mathbf{x},y)$ by substituting Eq~(\ref{eq:vi}) and Eq~(\ref{eq:vo}) into Eq~(\ref{eq:v}) and have
\begin{equation}
\begin{aligned}
& \widetilde{P}_I(\mathbf{x},y) = \frac{1}{N_I} \sum_{i = 1}^{N_I} \delta(\mathbf{x} = \mathbf{x}_i^I, y = y_i^I), \\
& \widetilde{P}_O(\mathbf{x},y) = \frac{1}{N_I} \sum_{i = 1}^{N_I} \mathbb{E}_{\mathbf{x}_1^I}\ldots\mathbb{E}_{\mathbf{x}_{M - 1}^I}\left[ \delta \left(\mathbf{x} = \mathbf{x}^O, y = y^O\right)\right].
\label{eq:v2}
\end{aligned}
\end{equation}
Both empirical distributions are estimated by in-distribution samples. The dirac delta vicinity distribution $\mathcal{VI}$ in $\widetilde{P}_I(\mathbf{x},y)$ degenerates into a simple Dirac delta function without exploring the samples outside the training dataset $\mathcal{D}$. Conversely, the cross-class vicinity distribution $\mathcal{VO}$ in $\widetilde{P}_O(\mathbf{x},y)$ combines different classes of in-distribution samples to find the out-of-distribution samples outside $\mathcal{D}$.

\subsection{Generic Empirical Risk}
In the pretrained phase, only in-distribution samples are available, with $P_I(\mathbf{x}, y) \approx \widetilde{P}_I(\mathbf{x},y)$, we approximate the expected risk $\mathcal{R}(\theta)$ by the following empirical risk
\begin{equation}
\mathcal{R}(\theta) \approx - \sum_{i = 1}^{N_I} \log Q_{\theta}(y_i^I | \mathbf{x}_i^I),
\label{eq:r1}
\end{equation}
and learn the pretrained network $Q_{\theta}$ by minimizing Eq. (\ref{eq:r1}) on in-distribution samples. To improve the out-of-distribution sensitivity of the pretrained network $Q_{\theta}$ in the finetuning phase, we consider a generic expected risk Eq.~(\ref{eq:ger}), which learns to reject the mapping between the inputs and the corresponding complementary labels by introducing out-of-distribution samples. Using $P_I(\mathbf{x}, y) \approx \widetilde{P}_I(\mathbf{x},y)$ and $P_O(\mathbf{x}, y) \approx \widetilde{P}_O(\mathbf{x},y)$, we approximate the generic expected risk $\mathcal{R}(\theta)$ by the following generic empirical risk:
\begin{equation}
\begin{aligned}
& \mathcal{R}(\theta) \approx \mathcal{\widetilde{R}}(\theta) \\
= & - \sum_{i = 1}^{N_I} \log Q_{\theta}(y_i^I | \mathbf{x}_i^I) - \sum_{j = 1}^{N_O} \log \left( 1 - Q_{\theta}(y_j^O |\mathbf{x}_j^O)\right),
\label{eq:r2}
\end{aligned}
\end{equation}
where $N_O$ is the number of out-of-distribution samples drawn from $\widetilde{P}_O(\mathbf{x},y)$. Based on Monte Carlo~\cite{MC:19}, we apply the stochastic gradient descent optimization algorithm~\cite{ML:14} to estimate the gradients of Eq. (\ref{eq:r1}) and Eq. (\ref{eq:r2}). The pseudo-code of the finetuning procedure is summarized in Algorithm~\ref{alg:pc}.

\section{Proof}~\label{sec:pf}
\subsection{Lower bound of $\mathcal{I}_I(\mathbf{x};y)$} \label{App:LWB}
For the mutual information $\mathcal{I}_I(\mathbf{x};y)$ measuring the relationship between inputs and labels of in-distribution samples drawn from the joint distribution $P_I(\mathbf{x},y)$, we have
\begin{equation}
\begin{aligned}
\mathcal{I}_I(\mathbf{x};y) = &  \mathbb{E}_{P_I(\mathbf{x},y)} \left[\log \frac{P_I(y | \mathbf{x})}{P_I(y)}\right] \\
= & \mathbb{E}_{P_I(\mathbf{x},y) } \left[\log \frac{P_I(y | \mathbf{x}) Q_{\theta}(y | \mathbf{x}) }{P_I(y) Q_{\theta}(y | \mathbf{x})}\right] \\
= & \mathbb{E}_{P_I(\mathbf{x},y) } \left[ \log Q_{\theta}(y | \mathbf{x}) \right] \\
& + D_{KL}\left( P_I(\mathbf{x},y) \| Q_{\theta}(y | \mathbf{x})\right) + H(P_I(y)) \\
\geq & \int \log Q_{\theta}(y | \mathbf{x}) {\rm d} P_I(\mathbf{x},y) + H(P_I(y))
\end{aligned}
\end{equation}
where the inequality is attributed to the nonnegative property of the Kullback-Leibler divergence.

\subsection{Upper bound of $\mathcal{I}_O(\mathbf{x};y)$}
\label{App:UWB}
For the mutual information $\mathcal{I}_O(\mathbf{x};y)$ measuring the relationship between inputs and complementary labels of out-of-distribution samples drawn from the joint distribution $P_O(\mathbf{x},y)$, we have
\begin{equation}
\begin{aligned}
\mathcal{I}_O(\mathbf{x};y) & = \mathbb{E}_{P_O(\mathbf{x},y)} \left[\log \frac{P_O(y | \mathbf{x})}{P_O(y)}\right] \\
& = \mathbb{E}_{P_O(\mathbf{x},y)} \left[\log \frac{P_O(y | \mathbf{x}) \left( 1 - Q_{\theta}(y | \mathbf{x})\right)}{P_O(y)\left( 1 - Q_{\theta}(y | \mathbf{x})\right)}\right]\\
\leq & - \mathbb{E}_{P_O(\mathbf{x},y) } \left[ \log \left( 1 - Q_{\theta}(y | \mathbf{x})\right) \right] \\
& + \mathbb{E}_{P_O(\mathbf{x},y) } \left[ \log \frac{\left( 1 - Q_{\theta}(y | \mathbf{x})\right)}{P_O(y)}\right] \\
\leq & - \mathbb{E}_{P_O(\mathbf{x},y) } \left[ \log \left( 1 - Q_{\theta}(y | \mathbf{x})\right) \right] - 1 \\
& + \mathbb{E}_{P_O(\mathbf{x},y) } \left[ \frac{1 - Q_{\theta}(y | \mathbf{x})}{P_I(y)}\right] + \mathbb{E}_{P_O(\mathbf{x},y) } \left[ \log \frac{P_I(y)}{P_O(y)} \right]  \\
\leq & - \mathbb{E}_{P_O(\mathbf{x},y) } \left[ \log \left( 1 - Q_{\theta}(y | \mathbf{x})\right) \right] \\
& + \mathbb{E}_{P_O(\mathbf{x},y) } \left[ \frac{1}{P_I(y)}\right] - D_{KL}\left( P_O(y) \| P_I(y)\right)\\
\leq & - \int \log (1 - Q_{\theta}(y | \mathbf{x})) {\rm d} P_O(\mathbf{x},y)  + \mathbb{E}_{P_O(y) } \left[ \frac{1}{P_I(y)}\right] \\
\end{aligned}
\end{equation}
where the first inequality is attributed to $P_O(y | \mathbf{x}) \geq 1$; the second inequality uses the logarithm inequality $\log(x) \leq \frac{x}{a} + \log(a) - 1$ for all $x,a \geq 0$; the third inequality is attributed to $Q_{\theta}(y | \mathbf{x}) \geq 0$; the last is attributed to the nonnegative property of the Kullback-Leibler divergence.

\subsection{Proof of Theorem~\ref{th:kc}}
\label{App:proof}
\begin{figure}
\subfigure[]{
    \begin{minipage}{0.47\linewidth}
    \centering
    \includegraphics[width=1\linewidth]{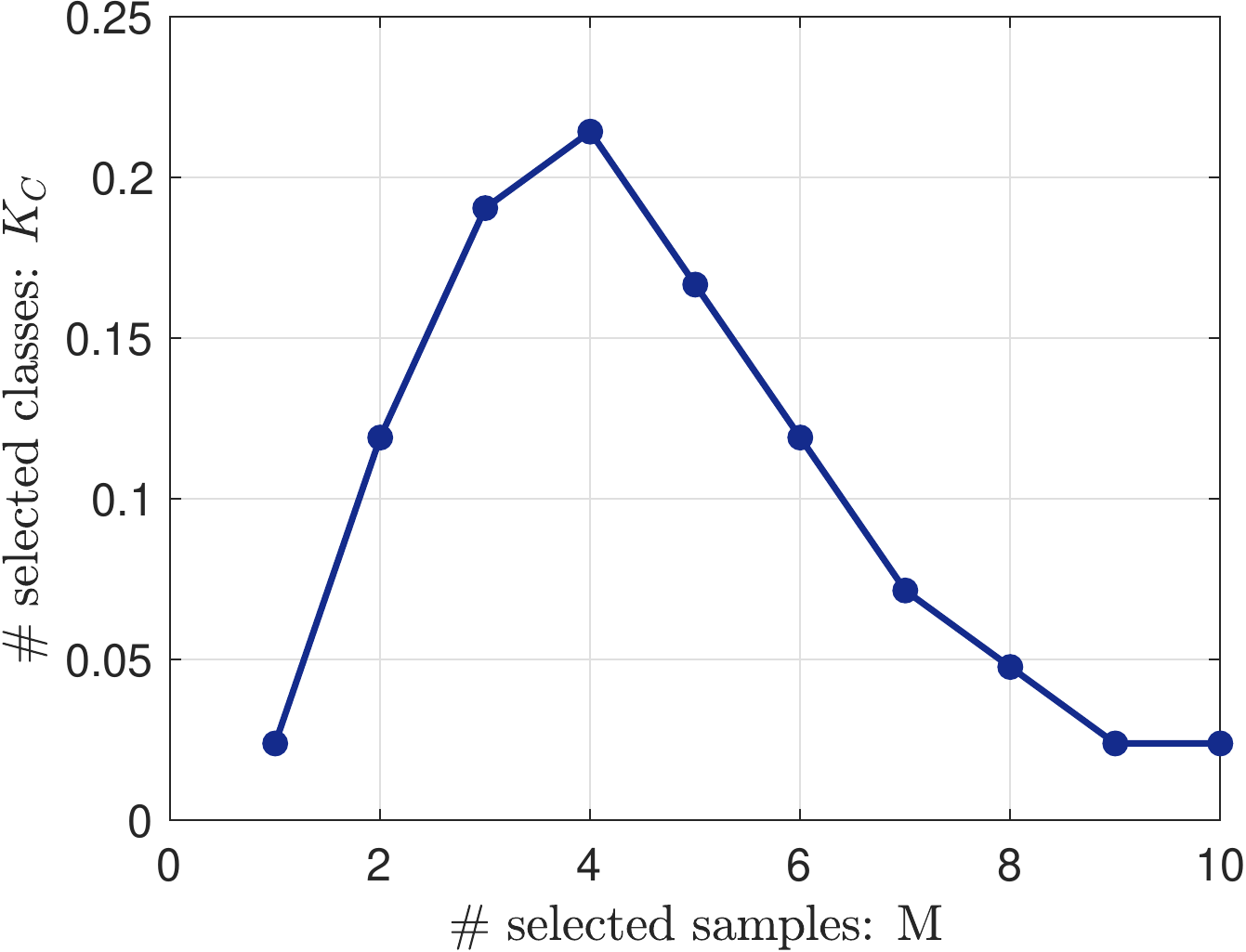}
    \vspace{0.01em}
    \end{minipage}
    \label{fig:prob1}
  }
\subfigure[]{
    \begin{minipage}{0.46\linewidth}
    \centering
    \includegraphics[width=1\linewidth]{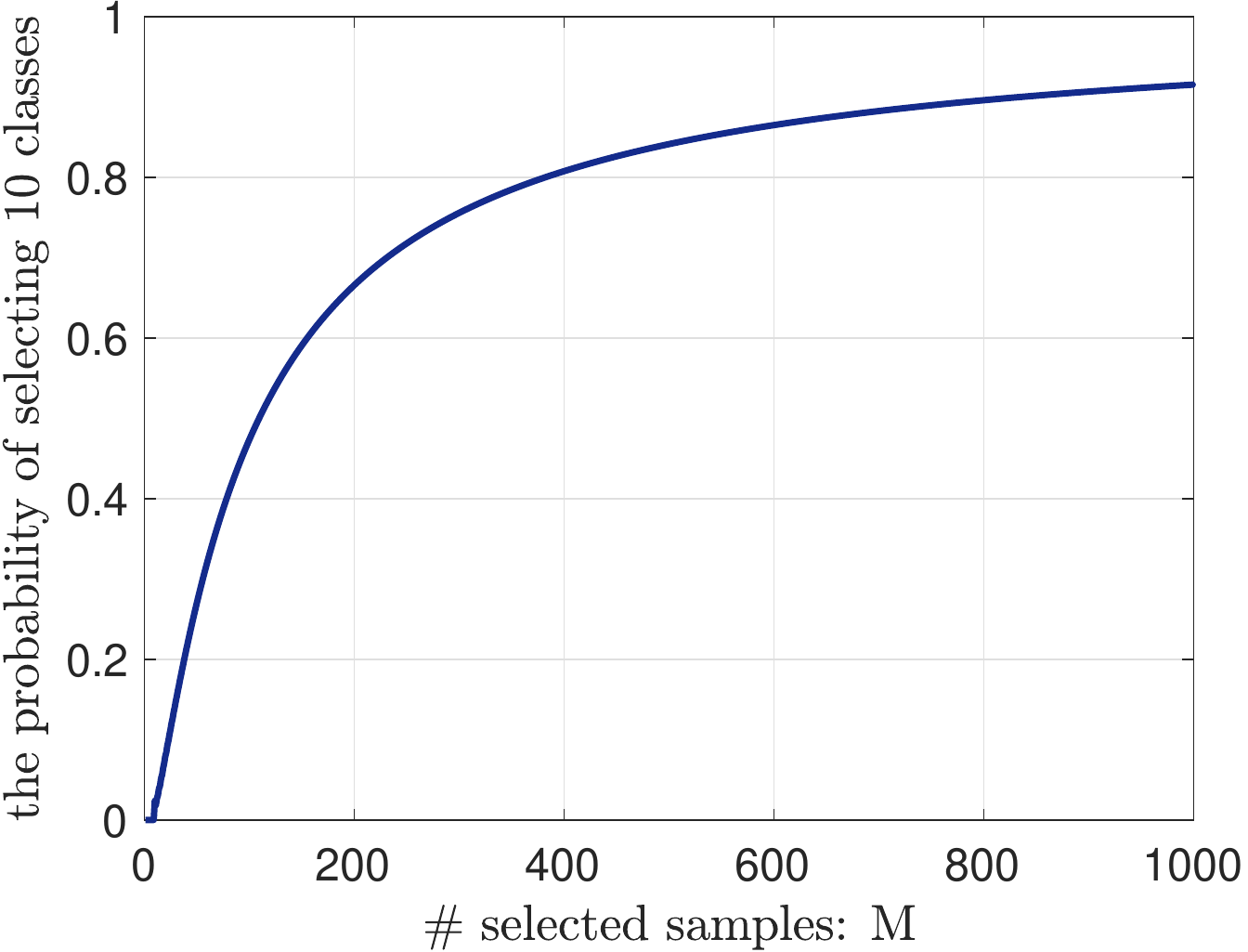}
    \vspace{0.01em}
    \end{minipage}
    \label{fig:prob2}
  }
  \caption{(a) An illustration of the distribution over $K_C$ when the number of classes $K$ and the number of selected samples $M$ are equal to $10$. (b) An illustration of the probability of $K_C = K = 10$ for the different number of selected samples $M$.}
\end{figure}

The problem of calculating the number of selected classes $K_C$ for $M$ selected in-distribution samples and $K$ classes is equivalent to calculating the number of allocation schemes for $M$ balls and $K$ boxes where the boxes could be empty~\cite{BB:21}. According to dynamic programming~\cite{DP:19}, the number of selected classes $K_C$ among the $M$ samples satisfies the following distribution,
\begin{equation}
\begin{aligned}
P(K_C) & = \frac{d(M, K_C)}{ \sum_{i = 1}^K d(M, i)} \\
\end{aligned}
\end{equation}
where
\begin{equation}
\begin{aligned}
d(M, K_C) & = \sum_{i = 1}^{K_C} d(M - K_C, i), M \geq K_C,\\
d(M, K_C) & = 0, M < K_C,\\
d(M,1) & = d(M,M) = 1.\\
\end{aligned}
\end{equation}
\figurename~\ref{fig:prob1} shows the distribution over $K_C$ when $M = K = 10$. When the number of components in $\mathbf{x}^O$ equals $K = 10$, the number of labels contained in $\mathbf{x}^O$ is $K_C (K_C \leq K)$, and the highest probability corresponds to $4$ different classes. The probability of $K_C = K = 10$ for the different number of selected samples $M$ is presented in \figurename~\ref{fig:prob2}. It indicates that the $M$ selected samples can only contain all classes of samples when $M$ is extremely large $M > 1000$.

\begin{algorithm}[h] \small
    \caption{The pseudo-code of LCVD.}
    \label{alg:pc}
    \begin{algorithmic}[1]
    \STATE {\textbf{Input:} pretrained network $Q_{\theta}$, \\
    \quad \quad \quad in-distribution training dataset $\mathcal{D} = \{(\mathbf{x}_i^I, y_i^I)\}_{i = 1}^{N_I}$, \\
    \quad \quad \quad batch size $b$, learning rate $\mu$}
    \REPEAT
    \STATE Draw $b_I = b / 2$ in-distribution samples from $\widetilde{P}_I(\mathbf{x},y)$: $\{(\mathbf{x}_i^I,y_i^I)\}_{i = 1}^{b_I}$
    \STATE Draw $b_O = b / 2$ out-of-distribution samples from $\widetilde{P}_O(\mathbf{x},y)$: $\{(\mathbf{x}_j^O, y_j^O)\}_{j = 1}^{b_O}$
    \STATE Estimate the objective function: \\
        \begin{align*}
        \mathcal{\widetilde{R}}(\theta) = - \sum_{i = 1}^{b_I} \log Q_{\theta}(y_i^I | \mathbf{x}_i^I) - \sum_{j = 1}^{b_O} \log \left( 1 - Q_{\theta}(y_j^O |\mathbf{x}_j^O)\right)
        \end{align*}
    \STATE Obtain gradients: $\nabla_{\theta} \mathcal{\widetilde{R}} (\theta)$
    \STATE Update parameters: $\theta  = \theta + \mu \nabla_{\theta} \mathcal{\widetilde{R}}(\theta)$
    \UNTIL{convergence}
    \STATE {\bfseries Output:} finetuned network $Q_{\theta}$
\end{algorithmic}
\end{algorithm}

\section{Experiments}~\label{sec:Experiments}
In this section, we verify the effectiveness of the proposed LCVD method. We compare it with different out-of-distribution detection methods in terms of out-of-distribution detection performance, in-distribution classification accuracy, and the efficiency of generating out-of-distribution samples. Furthermore, we analyze the effect of the number of selected in-distribution samples $M$ on constructing out-of-distribution samples, compare different training mechanisms with the generated out-of-distribution samples, and run a set of ablation study experiments about the inputs and labels of the generated out-of-distribution samples\footnote{The source codes are available at: \url{https://github.com/Lawliet-zzl/LCVD}.}.

\subsection{Setup}
We adopt the ResNet18 architecture~\cite{RES:16} for all the experiments and implement it in PyTorch. In the pretraining and retraining phase of all methods, the learning rate starts at $0.1$ and is divided by $10$ after $100$ and $150$ epochs, and all networks are trained for $200$ epochs with $128$ samples per mini-batch. In the finetuning phase of the proposed LCVD, the finetuning learning rate is $0.001$, which equals the final learning rate in the pretraining phase. If not specified, we randomly select $M = 10$ in-distribution samples to construct an out-of-distribution sample in the proposed method and finetune the pretrained network until convergence.

We can only access in-distribution datasets to train networks in the training phase. In the test phase, we evaluate the out-of-distribution detection performance on diverse real-world out-of-distribution datasets and the test in-distribution datasets corresponding to the training ones. The training in-distribution datasets used in our experiments to train neural networks include CIFAR10~\cite{CIFAR10:09}, SVHN~\cite{SVHN:11}, and Mini-Imagenet~\cite{IMAGENET:09}. The numbers of classes of CIFAR10 and SVHN are $10$, and the resolution ratios are $32 \times 32$ of the two in-distribution datasets. The number of classes and the resolution ratio are $10$ and $224 \times 224$ of Mini-Imagenet, respectively. For data augmentation methods, we apply random cropping and random horizontal flipping to CIFAR10 and SVHN and resizing and random cropping to Mini-Imagenet. To evaluate the detection performance, the test out-of-distribution datasets include CIFAR100~\cite{CIFAR10:09}, CUB200~\cite{CUB200}, StanfordDogs120~\cite{StanfordDogs120}, OxfordPets37~\cite{OxfordPets37}, Oxfordflowers102~\cite{Oxfordflowers102}, Caltech256~\cite{CAL:06}, DTD47~\cite{DTD47}, and COCO~\cite{COCO:14}. We resize the test out-of-distribution samples to match the size of training in-distribution samples.

To calculate the out-of-distribution scores of test in- and out-of-distribution samples, if not specified, we apply the threshold-based detector in the baseline method~\cite{BL:17} for the proposed LCVD method. An in-distribution sample is expected to have a high out-of-distribution score, while an out-of-distribution sample is the opposite. Following the setups of ODIN~\cite{ODIN:18}, to evaluate the out-of-distribution detection performance, we adopt the Area Under the Receiver Operating Characteristic curve (AUROC)~\cite{AUROC:06}, the Area under the Precision-Recall curve (AUPR), the false positive rate (FPR) at $95\%$ true positive rate (TPR), and the Detection Error to measure the rank of the out-of-distribution scores. AUPR includes AUPRIN and AUPROUT that present the area under the precision-recall curve where in- and out-of-distribution samples are treated as positives, respectively. Larger AUROC, AUPRIN, AUPROUT indicate better detection performance, while lower FPR at $95\%$ TPR, and Detection Error indicate better detection performance. If not specified, we report the average result across the abovementioned eight out-of-distribution datasets for each out-of-distribution detection method. The retraining and finetuning methods need to modify the given pretrained network. The retrained and finetuned networks sacrifice accuracy to improve the out-of-distribution sensitivity. We thus also evaluate the in-distribution classification accuracy to compare the price of improving out-of-distribution sensitivity. Furthermore, throughput~\cite{TP:21}, which represents the number of processed samples in a second, is adopted to measure the efficiency of generating out-of-distribution samples.

\subsection{Comparison Results}

\begin{table*}[t]
\renewcommand{\arraystretch}{1.3}
\setlength\tabcolsep{4pt}
\centering
\caption{The out-of-distribution detection performance of pretrained and finetuned network with diverse detectors. Each value represents the average AUROC across the eight out-of-distribution datasets. All values are in percentage, and boldface values show the relatively better detection performance.}
\label{tb:det}
\begin{tabular}{ccccccc}
\hline
\multirow{2}{*}{In-dist} & Baseline & ODIN & MLB  & Energy & RA   & GradNorm \\ \cline{2-7}
                         & \multicolumn{6}{c}{Pretrained / Finetuned (LCVD)}  \\ \hline \hline
CIFAR10                  & 78.7 / \textbf{82.4} & 79.5 / \textbf{84.6} & 79.9 / \textbf{82.5} & 77.8  / \textbf{82.5} & 79.8 / \textbf{83.1} & 80.3 / \textbf{83.6}\\ 
SVHN                     & 92.7 / \textbf{95.8} & 92.9 / \textbf{95.9} & 94.5 / \textbf{97.2} & 91.2  / \textbf{95.5} & 93.1 / \textbf{97.4} & 94.2 / \textbf{97.4}\\ 
Mini-Imagenet            & 75.1 / \textbf{78.6} & 75.4 / \textbf{81.2} & 73.1 / \textbf{76.6} & 74.2  / \textbf{77.4} & 78.5 / \textbf{82.1} & 76.1 / \textbf{81.5}\\ \hline
\end{tabular}
\end{table*}

\begin{table}[] \tiny
\renewcommand{\arraystretch}{1.3}
\setlength\tabcolsep{4pt}
\centering
\caption{The out-of-distribution detection performance of seven outlier exposure methods and LCVD. We calculate the average results of each comparison method across the eight test out-of-distribution datasets. Each value is obtained over five random trials of a method. The symbols $\uparrow$ and $\downarrow$ indicate a larger and lower value is better, respectively.}
\label{tb:comp}
\begin{tabular}{ccccccc}
\hline
In-dist                         & Method   & AUROC $\uparrow$  & AUPRIN $\uparrow$ & AUPROUT $\uparrow$ & FPR $\downarrow$ & Detection $\downarrow$\\ \hline \hline
\multirow{10}{*}{CIFAR10}       
                                & OE       & 78.1          & 77.1          & 76.4          & 73.1          & 26.3          \\ 
                                & POEM     & 79.5          & 73.7          & 78.2          & 67.2          & 26.7          \\ 
                                & MIXUP    & 76.5          & 74.3          & 77.3          & 65.3          & 27.2          \\ 
                                & JCL      & 78.1          & 79.8          & 77.6          & 74.3          & 25.7          \\ 
                                & JEM      & 74.3          & 68.2          & 74.8          & 72.2          & 27.9          \\ 
                                & CSI      & 79.2          & 75.7          & 78.4          & 67.8          & 24.5          \\ 
                                & SSD      & 80.2          & 76.7          & 79.4          & 66.8          & 24.5          \\ 
                                & LCVD     & \textbf{82.4} & \textbf{80.3} & \textbf{80.1} & \textbf{65.1} & \textbf{23.6} \\ \hline
\multirow{10}{*}{SVHN}          
                                & OE       & 92.1          & 90.0          & 92.4          & 38.7          & 16.1          \\ 
                                & POEM     & 94.5          & 94.2          & 92.8          & 32.4          & 12.0          \\ 
                                & MIXUP    & 92.9          & 90.3          & 93.4          & 36.3          & 11.6          \\ 
                                & JCL      & 93.1          & 91.7          & 92.6          & 32.6          & 15.9          \\ 
                                & JEM      & 93.7          & 91.4          & 91.2          & 36.8          & 16.5          \\ 
                                & CSI      & 93.9          & 91.0          & 92.0          & 39.3          & 11.7          \\ 
                                & SSD      & 94.4          & 92.1          & 93.0          & 38.3          & 11.6          \\ 
                                & LCVD     & \textbf{95.8} & \textbf{94.7} & \textbf{93.7} & \textbf{30.6} & \textbf{10.5} \\ \hline
\multirow{10}{*}{Mini-Imagenet} 
                                & OE       & 73.0          & 76.0          & 68.9          & 85.8          & 31.9          \\ 
                                & POEM     & 74.6          & 77.3          & 69.7          & 84.6          & 30.1          \\ 
                                & MIXUP    & 74.7          & 77.6          & 69.6          & 85.4          & 29.3          \\ 
                                & JCL      & 73.1          & 75.5          & 67.0          & 87.4          & 32.1          \\ 
                                & JEM      & 73.7          & 68.2          & 74.8          & 72.2          & 27.8          \\ 
                                & CSI      & 76.2          & 74.8          & 67.7          & 72.7          & 32.2          \\ 
                                & SSD      & 76.8          & 74.8          & 67.8          & 71.2          & 32.4          \\ 
                                & LCVD     & \textbf{78.6} & \textbf{79.6} & \textbf{76.7} & \textbf{70.7} & \textbf{20.4} \\ \hline
\end{tabular}
\end{table}

\begin{figure}
\centering
\subfigure{
    \begin{minipage}{0.46\linewidth}
    \centering
    \includegraphics[width=1\textwidth]{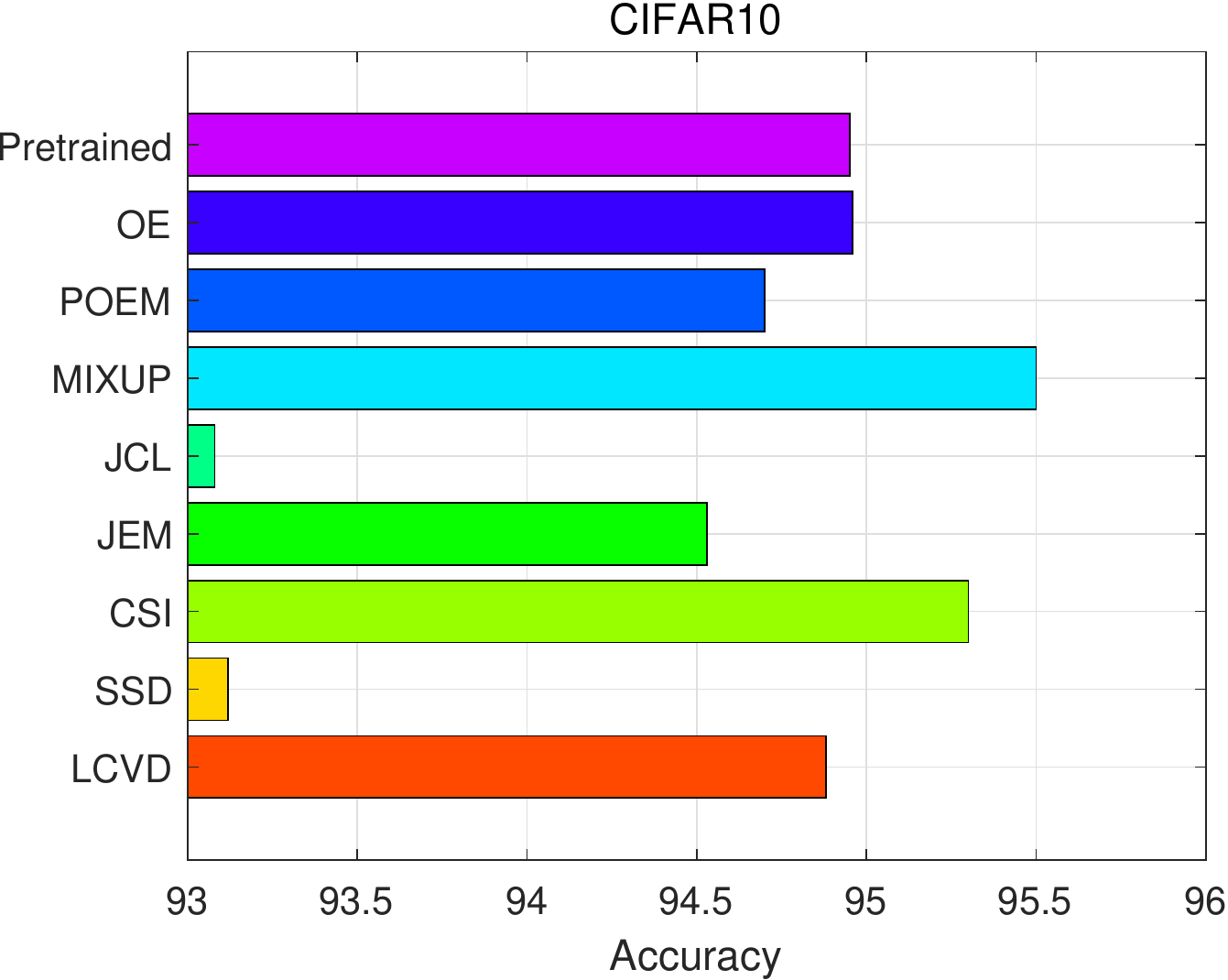}
    \end{minipage}
  }
\subfigure{
    \begin{minipage}{0.46\linewidth}
    \centering
    \includegraphics[width=1\textwidth]{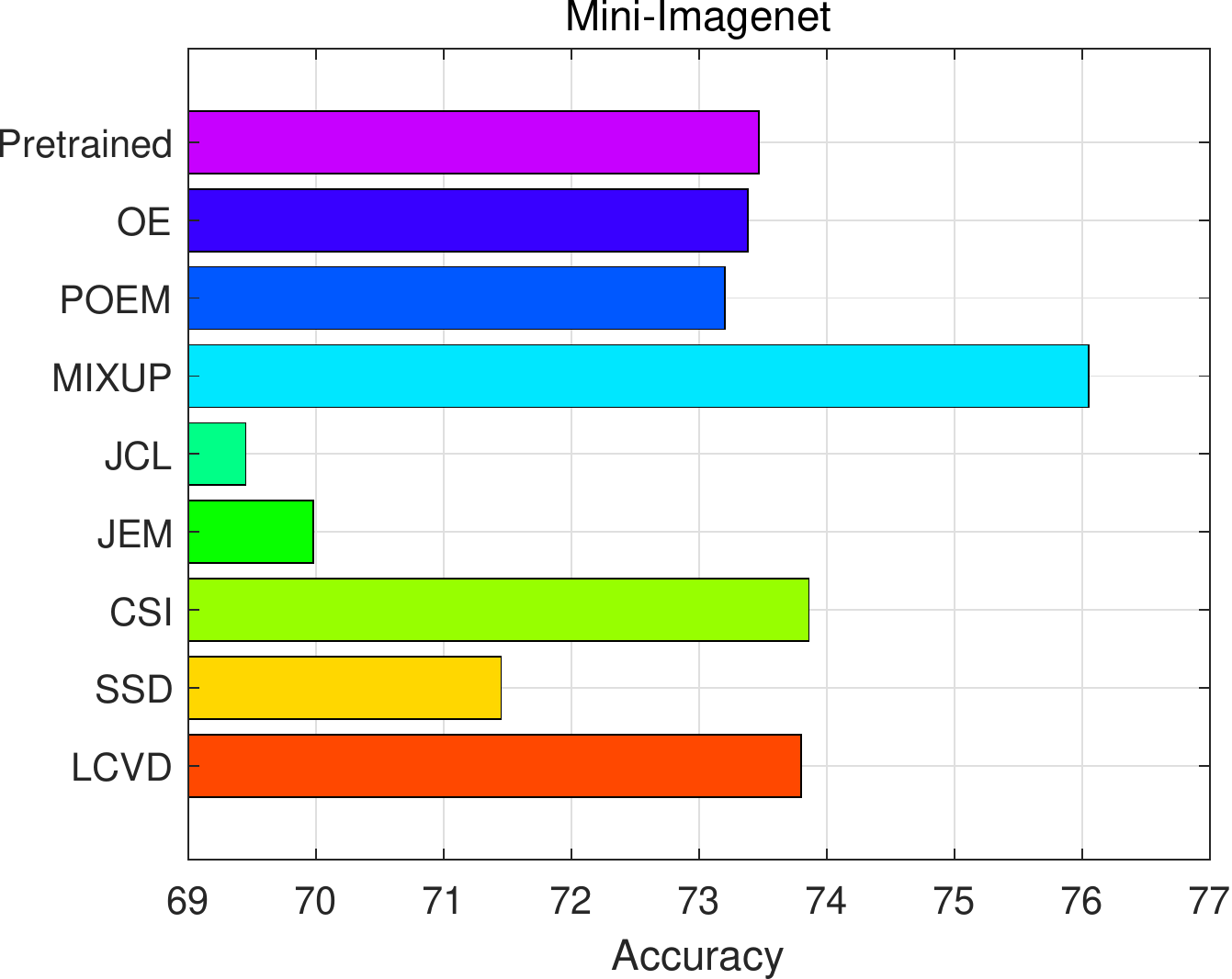}
    \end{minipage}
  }
  \caption{The in-distribution classification accuracy of the baseline, five retraining methods, and LCVD. A longer bar indicates a better classification.}
  \label{fig:acc}
\end{figure}

\subsubsection{Comparison with Different Detectors}
We incorporate six state-of-the-art out-of-distribution detectors into a pretrained network and its corresponding network finetuned by LCVD. The adopted detectors include the Baseline~\cite{BL:17}, ODIN~\cite{ODIN:18}, MahaLanoBis (MLB)~\cite{MLB:18}, energy-based detector (Energy)~\cite{EB:20}, Rectified Activations (RA)~\cite{RA:21}, and GradNorm~\cite{IG:21,ED:22}. For a fair comparison, we apply a modified MLB which adopts adversarial examples generated from in-distribution samples to train the logistic regression because the other compared methods do not need real-world out-of-distribution samples in the training phase. We follow the same setup as the original ones for the other comparison methods.


The results in \tablename~\ref{tb:det} show that the pretrained can achieve a significant improvement ($2.56 \%$ to $7.69 \%$) after being finetuned by LCVD for all out-of-distribution detectors. Specifically, LCVD leads to improved performance for the detectors applying softmax outputs, feature embeddings, and gradients. Therefore, LCVD can adapt different kinds of out-of-distribution detectors. This is mainly achieved by rejecting to map out-of-distribution inputs to their corresponding complementary labels. Specifically, LCVD draws data-dependent out-of-distribution samples with complementary labels from the cross-class vicinity distribution of in-distribution samples. Rejecting these complementary labels for the generated out-of-distribution samples in the finetuning processes improves the out-of-distribution sensitivity of pretrained networks.

\subsubsection{Comparison with Outlier Exposure Methods}
We compare the proposed LCVD method with seven outlier exposure methods that retrain or finetune a pretrained networks with out-of-distribution samples, namely Outlier Exposure (OE)~\cite{OE:19}, Posterior Sampling-based Outlier Mining (POEM)~\cite{POEM:22}, MIXUP~\cite{MIXUP:18}, Joint Confidence Loss (JCL)~\cite{GO:18}, Joint Energy-based Model (JEM)~\cite{EB:20}, Contrasting Shifted Instances (CSI)~\cite{CSI:20}, and Self-Supervised outlier Detection (SSD)~\cite{SSD:21}. OE and POEM require datasets to be with out-of-distribution samples, while the other methods generate out-of-distribution samples. For fair comparisons, following the suggestions of OE~\cite{OE:19}, we treat the in-distribution samples distorted with noise as out-of-distribution for OE and POEM. The settings of all the comparison methods follow the original ones. Each method is evaluated over five random trials.

The comparison results are summarized in \tablename~\ref{tb:comp}. LCVD achieves averagely $5.72 \%$, $2.45 \%$ and $5.42 \%$ improvement over the other state-of-the-art outlier exposure methods in terms of AUROC on datasets CIFAR10, SVHN, and Mini-Imagenet, respectively, which indicates that LCVD obtains the best performance. Furthermore, for the other metrics, LCVD also achieves the best out-of-distribution detection methods. The results demonstrate that the out-of-distribution samples drawn from the cross-class vicinity distribution of training in-distribution samples can effectively improve the out-of-distribution sensitivity of the pretrained network learned from the training in-distribution dataset. It is because the generated out-of-distribution samples are specific to the training in-distribution samples, which can explore the outside ranges of the training samples. The pretrained networks are encouraged to provide low-confidence predictions for samples in the outside ranges by refusing to map data-dependent out-of-distribution inputs to the corresponding complementary labels.

\begin{table*}
\renewcommand{\arraystretch}{1.3}
\setlength\tabcolsep{4pt}
\centering
\caption{The detection performance on near and far out-of-distribution samples in terms of AUROC. All values are in percentage, and boldface values show the relatively better detection performance.}
\label{tb:near}
\begin{tabular}{cccccccccc}
\hline
In-dist                  & Out-of-dist     & OE   & POEM & MIXUP & JCL  & JEM  & CSI   & SSD  & LCVD           \\ \hline \hline
\multirow{2}{*}{CIFAR10} & CIFAR100 (Near) & 88.7 & 88.9 & 84.4  & 84.5 & 82.9 & 88.9  & 89.3 & \textbf{89.8} \\ 
                         & SVHN (Far)      & 95.8 & 96.3 & 92.7  & 87.8 & 91.2 & 95.3  & 96.8 & \textbf{98.2} \\ \hline
\end{tabular}
\end{table*}

\begin{table*}
\renewcommand{\arraystretch}{1.3}
\setlength\tabcolsep{4pt}
\centering
\caption{The efficiency of generating out-of-distribution samples. Each value presents the throughput (image / s).}
\label{tb:eff}
\begin{tabular}{cccccccc}
\hline
In-dist       & Image Size & OE \& POEM & MIXUP  & JCL  & JEM & CSI \& SSD & LCVD  \\ \hline \hline
CIFAR10       & 32 $\times$ 32    & 29412      & 500000 & 1577 & 158 & 12500      & 71429 \\ 
Mini-ImageNet & 224 $\times$ 224  & 612        & 100000 & 382  & 56  & 3533       & 5814  \\ \hline
\end{tabular}
\end{table*}

\subsubsection{Comparison on Near and Far Samples}
Following the setups of Winkens et al.~\cite{CT:20}, we compare the detection performance of exposure methods on near and far samples. Near and far samples represent the out-of-distribution samples with covariate shift (slightly different from in-distribution samples) and semantic shift (significantly different from in-distribution samples), respectively. For the in-distribution dataset CIFAR10, SVHN can be treated as a far out-of-distribution dataset because the classes of the two datasets are vastly different. CIFAR100 can be treated as near out-of-distribution dataset because it contains some classes having the similar semantics with that of CIFAR10.

The results are reported in~\tablename~\ref{tb:near}. For the far out-of-distribution dataset (SVHN), all the methods achieve promising results, and LCVD achieves the best ($3.59 \%$ improvement). This is because the semantics of the two datasets are vastly different, which indicates that networks can easily separate the two kinds of samples. Accordingly, it is challenging to distinguish out-of-distribution samples from in-distribution samples when the two datasets have classes with similar semantics. Therefore, for the near out-of-distribution dataset (CIFAR100), the detection performance of all the methods decreases. However, LCVD still obtains the best detection performance ($4.91 \%$ improvement). This is because LCVD explores out-of-distribution samples by considering the vicinity relations between in-distribution samples of different classes, and some constructed out-of-distribution samples are near. Learning to reject these constructed out-of-distribution inputs to their corresponding complementary labels in the finetuning process, a pretrained network is encouraged to provide low-confidence predictions for out-of-distribution samples.

\subsubsection{Comparison of Classification Accuracy}
The comparison results in terms of classification accuracy are summarized in Fig.~\ref{fig:acc}. We observe that only the CSI and MIXUP methods outperform the pretrained networks on the two datasets. This is because the target of MIXUP is to improve the generalization rather than out-of-distribution detection by generating more in-distribution samples by convex combinations, and the rotated samples used in CSI are also in-distribution. The augmented in-distribution samples are more beneficial to learn invariable features to improve the performance of recognizing different classes of in-distribution samples. We can also observe that the classification performance of LCVD is close to the baseline method on CIFAR10 and slightly better on Mini-Imagenet. However, the rest methods obtain poor classification performance. Therefore, LCVD can significantly improve the out-of-distribution sensitivity by sacrificing only tiny classification accuracy. It is because LCVD generates out-of-distribution samples by augmenting the training in-distribution samples, and the adaptively generated samples make a less negative impact on the classification learning process.

\subsubsection{Efficiency Analysis}
We measure the efficiency of outlier exposure methods in generating out-of-distribution samples in terms of throughput. The results are summarized in \tablename~\ref{tb:eff}. We observe that MIXUP is the most effective method because it convexly combines two randomly selected in-distribution inputs to construct an out-of-distribution input without allocating extra memory. The proposed LCVD is the second effective method. Specifically, compared with the other outlier exposure methods, LCVD efficiently generates the out-of-distribution samples for Mini-Imagenet, which indicates that LCVD is applicable for high-resolution samples. This is because LCVD generates out-of-distribution samples by augmenting in-distribution samples without training extra generators. Specifically, it linearly combines multiple randomly selected in-distribution inputs to construct an out-of-distribution input without training, which can be treated as an extension of MIXUP.

\begin{figure}[t]
  \centering
  \includegraphics[width=1\linewidth]{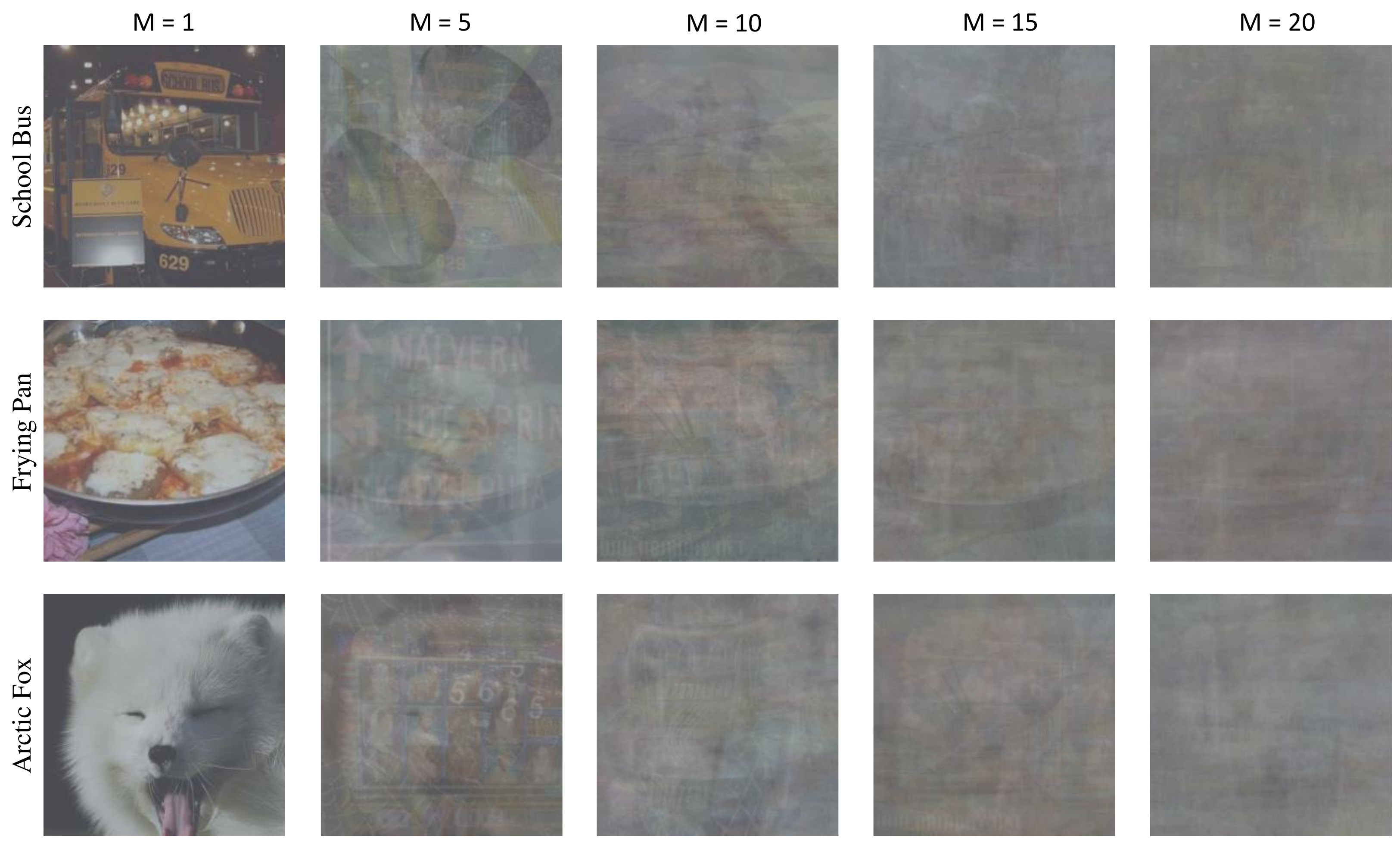}\\
  \caption{The out-of-distribution samples drawn from the cross-class vicinity distribution of the training in-distribution samples in Mini-Imagenet.}
  \label{fig:OOD}
\end{figure}

\begin{figure*}
\centering
\includegraphics[width=1\textwidth]{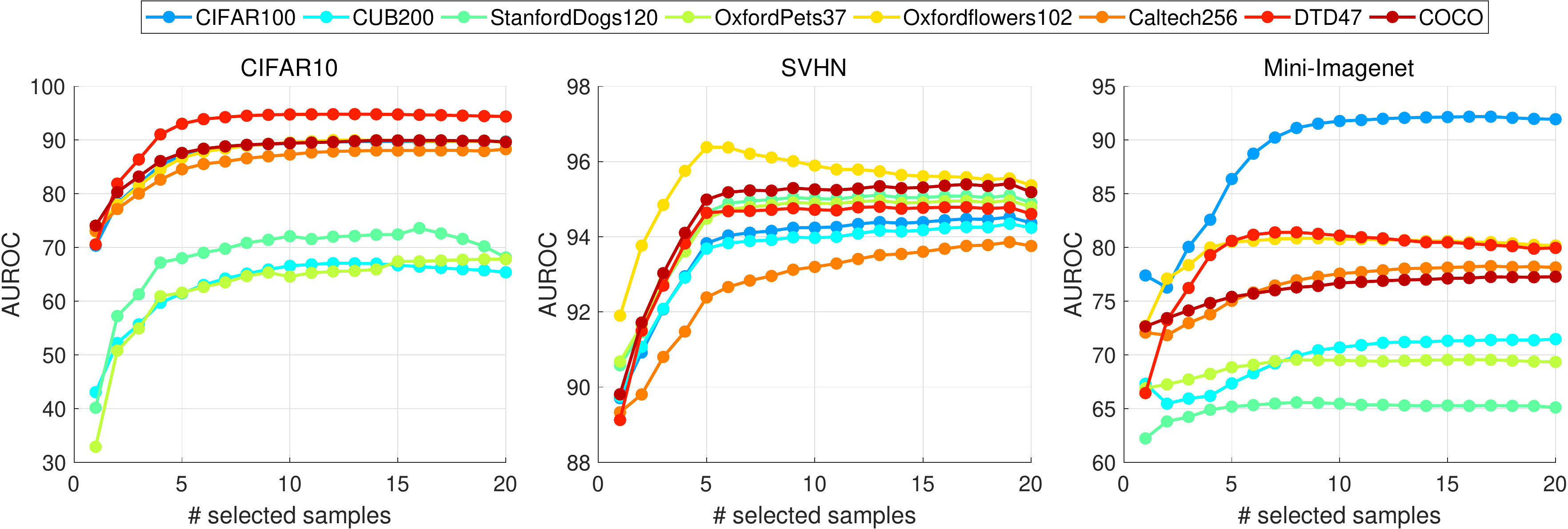}
  \caption{The effect of the number of selected in-distribution samples $M$ for constructing an out-of-distribution sample. Each point indicates an AUROC value, and each line represents an out-of-distribution dataset.}
  \label{fig:para}
\end{figure*}

\subsection{Parameter Analysis}
We analyze the effect of the number of selected in-distribution samples $M$ on constructing an out-of-distribution sample. $M$ is from $\{1, 5,10,15,20\}$. According to \figurename~\ref{fig:prob1} and \figurename~\ref{fig:prob2}, we know that the selected in-distribution samples cannot cover all labels with a high probability even if $M = 20$ on the three training in-distribution datasets.

The generated out-of-distribution samples with different $M$ are presented in \figurename~\ref{fig:OOD}. Note that, when $M = 1$, the original training in-distribution samples are treated as out-of-distribution samples. When $M = 5$, we can observe that the generated samples are still similar to the corresponding in-distribution samples, which indicates the generated samples tend to be in-distribution. When $M \geq 10$, the generated samples are significantly different from the corresponding in-distribution samples, which indicates the generated samples tend to be out-of-distribution. Furthermore, different in-distribution samples lead to diverse generated out-of-distribution samples.

The experimental results are shown in Fig.~\ref{fig:para}. We observe that the out-of-distribution detection performance increases with the increase of $M$, although this trend diminishes when $M$ is sufficiently large (e.g. $M \geq 20$). Combining many in-distribution samples to construct an out-of-distribution sample is expensive. Therefore, we apply $M = 10$ for LCVD which can balance efficiency and effectiveness. When $M$ is small, the generated samples tend to be in-distribution. The pretrained network rejecting these samples reduces the prediction confidence for in-distribution samples, which narrows the confidence gap between in- and out-of-distribution samples. Conversely, when $M$ is large, the generated samples tend to be out-of-distribution. The pretrained network rejecting these samples increases the prediction confidence on in-distribution samples, which enlarges the confidence gap between in- and out-of-distribution samples.

\subsection{Training mechanism}
\begin{table}[] \small
\renewcommand{\arraystretch}{1.3}
\setlength\tabcolsep{4pt}
\centering
\caption{The effect of the retraining and finetuning mechanisms. The results are the average AUROC value across the eight test out-of-distribution datasets. All values are in percentage, and the boldface values show the relatively better detection performance.}
\label{fig:tra}
\begin{tabular}{ccc}
\hline
In-distribution & Retrain & Finetune \\ \hline \hline
CIFAR10         & 81.7   & \textbf{82.4}    \\ 
SVHN            & 94.4   & \textbf{95.8}    \\ 
Mini-Imagenet   & 76.8   & \textbf{78.6}    \\ \hline
\end{tabular}
\end{table}

We analyze the effect of the two different training mechanisms, i.e., retraining and finetuning, on improving the out-of-distribution sensitivity of pretrained networks with the out-of-distribution samples drawn from the cross-class vicinity distribution of training in-distribution samples. The results are presented in \tablename~\ref{fig:tra}. We observe that the performance gaps between the two different mechanisms are narrow, and the finetuning method is better ($0.85 \%$ to $2.34 \%$) than the retraining method. Both the pretrained network and the generated out-of-distribution samples depend on the training in-distribution dataset. To reject those out-of-distribution samples, the network is better to have the acquired knowledge about the in-distribution samples. Based on the learned knowledge, the pretrained network can better discriminate between the out-of-distribution samples. Otherwise, the retrained network decreases the prediction confidence in in-distribution samples to narrow the confidence gap between in- and out-of-distribution samples. This is because the out-of-distribution samples generated by mixing multiple in-distribution samples still contain the in-distribution information. Considering both effectiveness and efficiency, we apply the finetuning mechanism for LCVD when a pretrained network is given.

\subsection{Ablation Study}
We run a set of ablation study experiments to verify that the input linearly combining multiple in-distribution samples and the corresponding complementary labels are indispensable for generating effective out-of-distribution samples.

\subsubsection{Diverse Out-of-distribution Inputs}
For the generated out-of-distribution samples in LCVD, we replace the inputs with the Gaussian noise and the rotation of in-distribution inputs~\cite{SSL:19}, respectively, without changing the complementary labels to obtain the other two variants. The Gaussian noise and rotation inputs correspond to complementary labels. We thus apply the same objective function Eq.~(\ref{eq:r2}) as for LCVD to refine the pretrained network on the two variants. The comparison results of diverse out-of-distribution inputs are presented in Fig.~\ref{fig:abla1}. We observe that rotation obtains the worst result because the rotated inputs are still essentially in-distribution, which should not be rejected by the pretrained network. Gaussian achieves better results because Gaussian noise inputs are out-of-distribution. However, Gaussian noise inputs are independent of the training in-distribution samples. That is the reason why LCVD considers the relations among the samples of different classes to generate specific out-of-distribution samples and achieves the best result.

\subsubsection{Diverse Out-of-distribution Labels}
\begin{figure}
\centering
\subfigure[Effect of Out-of-distribution Input]{
    \begin{minipage}{1\linewidth}
    \centering
    \includegraphics[width=1\textwidth]{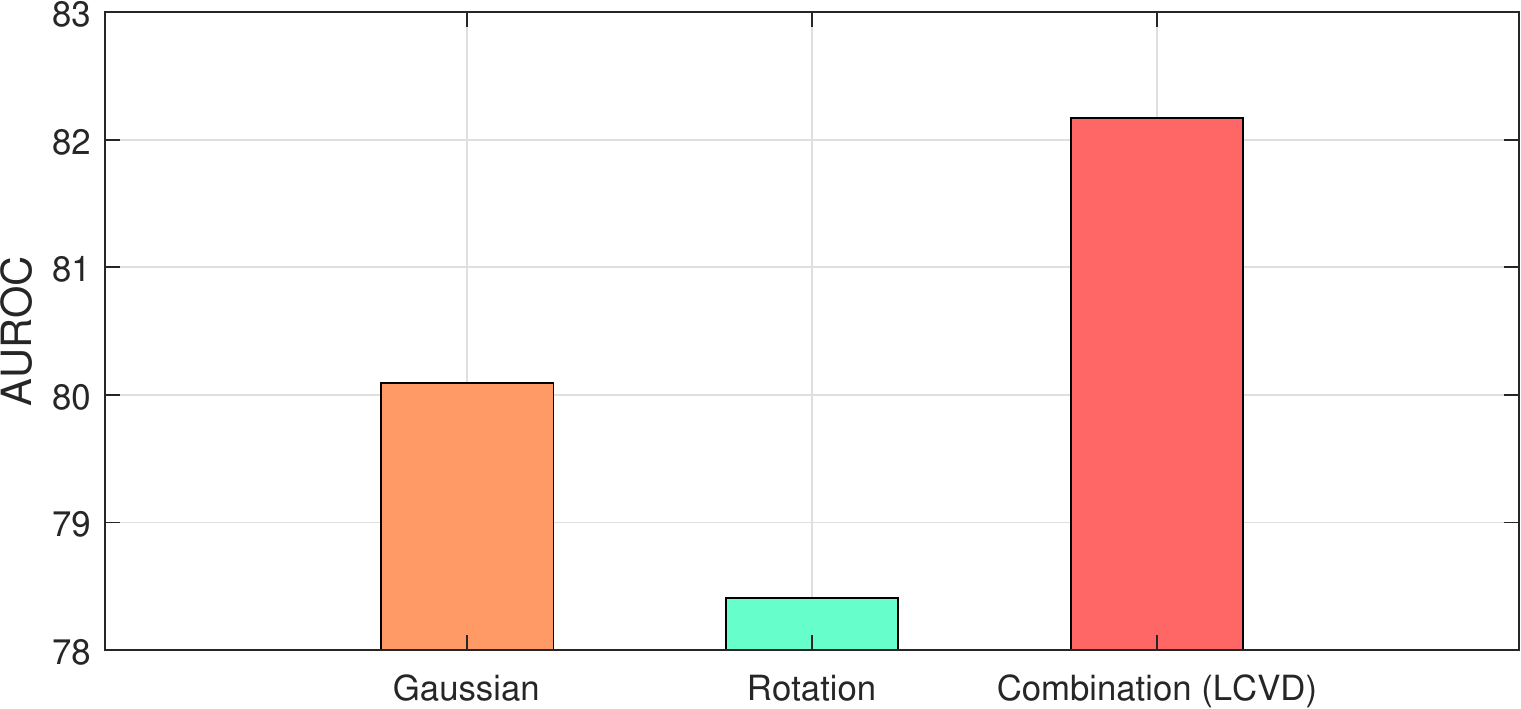}
    \vspace{0.05em}
    \end{minipage}
    \label{fig:abla1}
  }
\subfigure[Effect of Out-of-distribution Labels]{
    \begin{minipage}{1\linewidth}
    \centering
    \includegraphics[width=1\textwidth]{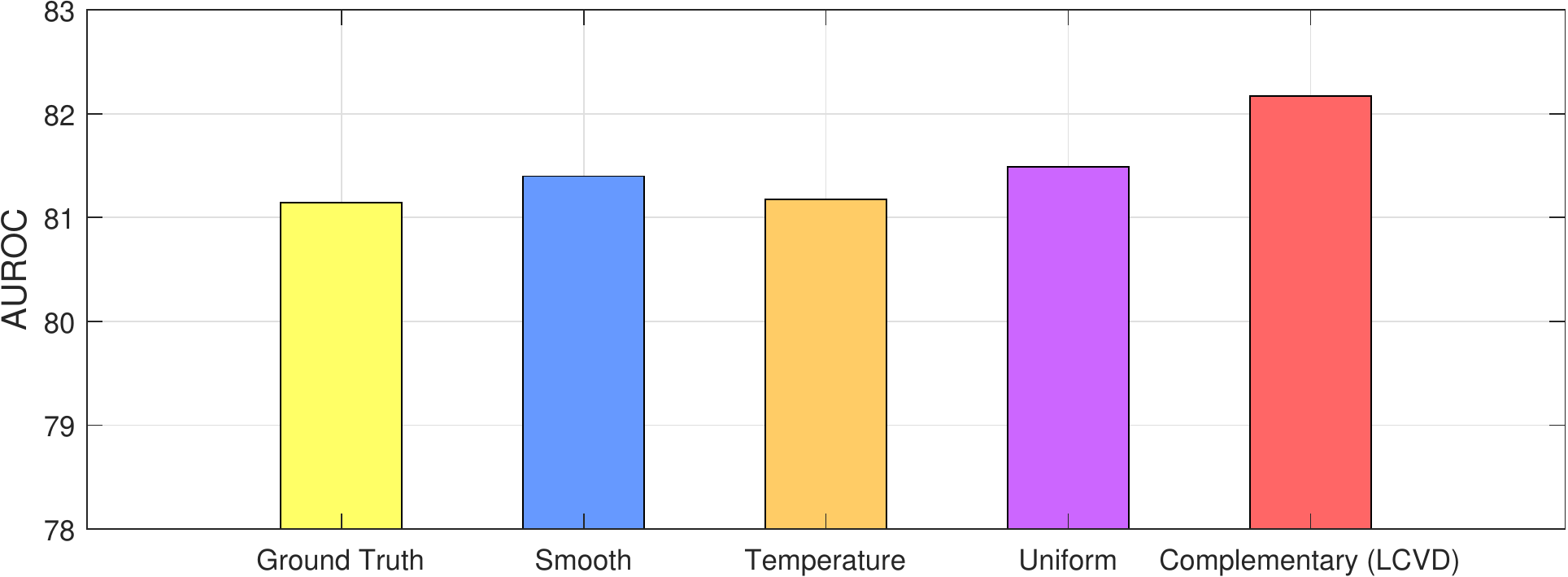}
    \vspace{0.05em}
    \end{minipage}
    \label{fig:abla2}
  }
  \caption{Results of the ablation study on the CIFAR10 dataset. (a) Replacing the inputs (that linearly combine multiple in-distribution inputs) of the constructed out-of-distribution samples in LCVD with other kinds of inputs. (b) Replacing the (complementary) labels of the constructed out-of-distribution samples in LCVD with other kinds of labels. Each bar indicates the average AUROC value across the eight test out-of-distribution datasets. A higher bar indicates a better detection result.}
  \vspace{-10pt}
\end{figure}

For the generated out-of-distribution samples in LCVD, without changing the inputs, we replace the complementary labels with four pseudo labels to obtain other four variants, respectively. The four pseudo labels include in-distribution Ground Truth labels, Smooth ground truth labels~\cite{CP:17}, ground truth labels with Temperature~\cite{KD:15}, and randomly selected label probability vectors from a Uniform distribution. The inputs of these four variants correspond to pseudo labels rather than complementary labels. We thus apply the objective function Eq.~(\ref{eq:r1}) used in the pretraining phase rather than the same objective function Eq.~(\ref{eq:r2}) as for LCVD to refine the pretrained network on the four variants. The comparison results of diverse out-of-distribution inputs are presented in Fig.~\ref{fig:abla2}. We can observe that LCVD achieves $1 \%$ improvement over the other generated out-of-distribution samples. This is because the other methods directly define the ground truth labels for the out-of-distribution inputs. Due to the complexity of inputs, the defined ground truth labels cannot precisely match the out-of-distribution inputs. Conversely, LCVD defines the complementary labels for the out-of-distribution inputs. It is hard to decide the ground truth labels. However, according to the construction method of the cross-class vicinity distribution, it is easy to determine which classes a generated out-of-distribution input does not belong to. Therefore, LCVD indirectly defines the learning targets for out-of-distribution inputs, which is more conservative and achieves the best result.

\section{Conclusion}
\label{sec:Conclusions}
In this paper, we propose the Learning from Cross-class Vicinity Distribution (LCVD) method which makes the first attempt to generate out-of-distribution samples by augmenting in-distribution samples. The cross-class vicinity distribution of in-distribution samples explores out-of-distribution samples by considering the vicinity relations between samples of different classes. An out-of-distribution input is generated by linearly combining multiple in-distribution inputs, which corresponds to a complementary label different from those labels of the constituent in-distribution samples. Given a pretrained network, we then finetune it to reject such out-of-distribution samples drawn from the cross-class vicinity distribution. This improves the out-of-distribution sensitivity of the pretrained network. Experiments show that LCVD makes better out-of-distribution detection than the state-of-the-art methods on diverse in- and out-of-distribution datasets. We further speculate that out-of-distribution samples are sensitive to neural architectures since different architectures lead to their respective output distributions. Accordingly, an interesting future direction is to generate out-of-distribution samples dependent on a pretrained network rather than a training in-distribution dataset. Furthermore, out-of-detection detection is highly related to predictive uncertainty and adversarial example detection. Accordingly, another interesting future direction is to explore augmented in-distribution samples for these research fields.


%

\ifCLASSOPTIONcaptionsoff
  \newpage
\fi

\bibliographystyle{IEEEtran}
\bibliography{refda}

\vspace{-0.7in}
\begin{IEEEbiography}[{\includegraphics[width=1in,height=1.25in,clip,keepaspectratio]{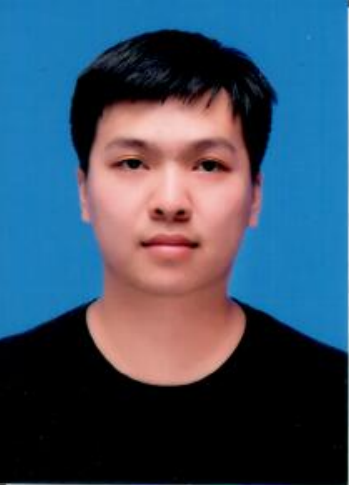}}]{Zhilin Zhao}
received a B.S. and M.S. degree from the School of Data and Computer Science, Sun Yat-Sen University, China. He is currently a PhD student in the School of Computer Science, University of Technology Sydney, Australia. His research interests include generalization analysis, online learning, and out-of-distribution detection.
\end{IEEEbiography}

\vspace{-0.7in}

\begin{IEEEbiography}[{\includegraphics[width=1in,height=1.25in,clip,keepaspectratio]{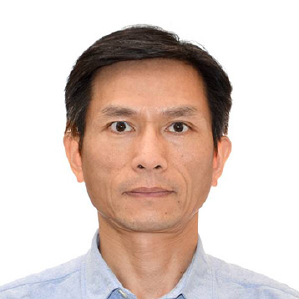}}]{Longbing Cao} 
is a Professor at the University of Technology Sydney and an ARC Future Fellow (Level 3). He received one PhD in Pattern Recognition and Intelligent Systems from the Chinese Academy of Sciences and another in Computing Science at UTS. His research interests include artificial intelligence, data science, knowledge discovery, machine learning, behavior informatics, complex intelligent systems, and enterprise innovation.
\end{IEEEbiography}

\vspace{-0.7in}

\begin{IEEEbiography}[{\includegraphics[width=1in,height=1.25in,clip,keepaspectratio]{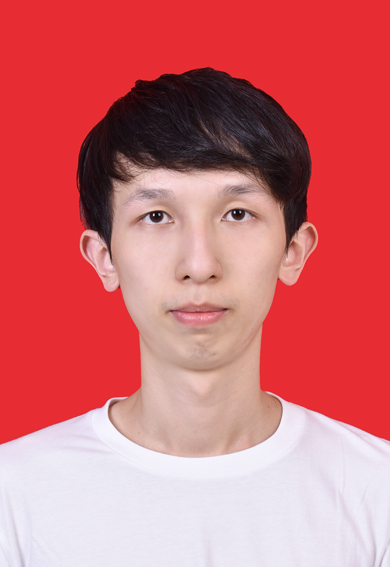}}]{Kun-Yu Lin}
received a B.S. and M.S. degree from the School of Data and Computer Science, Sun Yat-sen University, China. He is currently a PhD student in the School of Computer Science and Engineering, Sun Yat-sen University. His research interests include computer vision and machine learning.
\end{IEEEbiography}

\end{document}